\documentclass{article}

\usepackage{microtype}
\usepackage{graphicx}
\usepackage{subfigure}
\usepackage{booktabs} 

\usepackage{hyperref}

\usepackage[accepted]{icml2020}

\usepackage{subfigure}
\usepackage{multirow}
\usepackage{comment}
\usepackage{amsmath}        

\icmltitlerunning{Self-Supervised GAN Compression}

\begin{document}


\twocolumn[
\icmltitle{Self-Supervised GAN Compression}




\begin{icmlauthorlist}
\icmlauthor{Chong Yu}and \icmlauthor{Jeff Pool}
\begin{center}
NVIDIA \\
\{chongy,jpool\}@nvidia.com
\end{center}
\end{icmlauthorlist}




\vskip 0.4in
]




\begin{abstract}
Deep learning's success has led to larger and larger models to handle more and more complex tasks; trained models can contain millions of parameters.
These large models are compute- and memory-intensive, which makes it a challenge to deploy them with minimized latency, throughput, and storage requirements.
Some model compression methods have been successfully applied to image classification and detection or language models, but there has been very little work compressing generative adversarial networks (GANs) performing complex tasks.
In this paper, we show that a standard model compression technique, weight pruning, cannot be applied to GANs using existing methods.
We then develop a self-supervised compression technique which uses the trained discriminator to supervise the training of a compressed generator.
We show that this framework has a compelling performance to high degrees of sparsity, can be easily applied to new tasks and models, and enables meaningful comparisons between different pruning granularities.
\end{abstract}

\section{Introduction}

Deep Neural Networks (DNNs) have proved successful in various tasks like computer vision, natural language processing, recommendation systems, and autonomous driving. Modern networks are comprised of millions of para\-meters, requiring significant storage and computational effort. Though accelerators such as GPUs make realtime performance more accessible, compressing networks for faster inference and simpler deployment is an active area of research. Compression techniques have been applied to many networks to reduce memory requirements and improve performance. Though these approaches do not always harm accuracy, aggressive compression can adversely affect the behavior of the network. Distillation~\citep{schmidhuber1991neural,hinton2015distilling} can improve the accuracy of a compressed network by using information from the original, uncompressed network.

Generative Adversarial Networks (GANs)~\citep{schmidhuber1990making,goodfellow2014generative} are a class of DNN that consist of two sub-networks: a generative model and a discriminative model. Their training process aims to achieve a Nash Equilibrium between these two sub-models. GANs have been used in semi-supervised and unsupervised learning areas, such as fake dataset synthesis~\citep{radford2015unsupervised,brock2018large}, style transfer~\citep{zhu2017toward,azadi2018multi}, and image-to-image translation~\citep{zhu2017unpaired,choi2018stargan}. As with networks used in other tasks, GANs have millions of parameters and nontrivial computational requirements.

In this work, we explore compressing the generative model of GANs for efficient deployment. We show that applying standard pruning techniques causes the generator's behavior to no longer achieve the network's goal and that past work targeted at compressing GANs for simple image synthesis fall short when they are applied to pruning large tasks. In some cases, this result is masked by loss curves that look identical to the original training. By modifying the loss function with a novel combination of the pre-trained discriminator and the original and compressed generators, we overcome this behavioral degradation and achieve compelling compression rates with little change in the quality of the compressed generator's ouput. We apply our technique to several networks and tasks to show generality. Finally, we study the behavior of compressed generators when pruned with different amounts and types of sparsity, finding that a technique commonly used for accelerating image classification networks is not trivially applicable to GANs.

Our main contributions are:
\begin{itemize}
\vspace{-0.5em}
\setlength{\itemsep}{1pt}
\setlength{\parsep}{1pt}
\setlength{\parskip}{1pt}
    \item We illustrate that and explain why pruning the generator of a GAN with existing methods is unsatisfactory for complex tasks. (Section~\ref{sec-naiveGANCompression})
    \item We propose self-supervised compression for the generator in a GAN. (Section~\ref{sec-self-supervisedGeneratorCompression})
    \item We show that our technique can apply to several networks and tasks. (Section~\ref{sec-generalizationExperiments})
    \item We show and analyze qualitative differences in pruning ratio and granularities. (Section~\ref{sec-pruningRatioGranularity})
\vspace{-0.5em}
\end{itemize}

We leave performance gained by pruning GANs for future work, as it is dependent on the target hardware: there are various way to exploit fine-grained sparsity on CPUs \citep{elsen2019fast}, GPUs \citep{chen2018escoin,zhu2018sparse}, and custom accelerators \citep{SCNN2017,judd2017cnvlutin2}.


\section{Related Research}\label{sec-relatedResearch}

A common method of DNN compression is network pruning~\citep{han2015learning}: setting the small weights of a trained network to zero and fine-tuning the remaining weights to recover accuracy. \citet{zhu2017prune} proposed a gradual pruning technique (AGP) to compress the model during the initial training process.
\citet{wen2016learning} proposed a structured sparsity learning method that uses group regularization to force weights towards zero, leading to pruning groups of weights together. \citet{li2016pruning} pruned entire filters and their connecting feature maps from models, allowing the smaller network to be accelerated with standard dense software libraries. Though it was initially applied to image classification networks, network pruning has been extended to natural language processing tasks~\citep{See2016Compression,narang2017exploring} and to recurrent neural networks (RNNs) of all types - vanilla RNNs, GRUs~\citep{cho2014learning}, and LSTMs~\citep{hochreiter1997long}. As with classification networks, structured sparsity within recurrent units has been exploited~\citep{wen2017learning}.

A complementary method of network compression is quantization. Sharing weight values among a collection of similar weights by hashing~\citep{chen2015hashing} or clustering~\citep{han2016deep} can save storage and bandwidth at runtime. Changing fundamental data types affords hardware the ability to accelerate the arithmetic operations, both in training~\citep{paulius2018mixed} and inference regimes~\citep{jain2019quantization}.

Several techniques have been devised to combat lost accuracy due to compression, since there is always the chance that the behavior of the network may change in undesirable ways when the network is compressed. Using GANs to generate unique training data~\citep{liu2018model} and extracting knowledge from an uncompressed network, known as distillation~\citep{hinton2015distilling}, can help keep accuracy high. Since the pruning process involves many hyperparameters, ~\citet{lin2019towards} use a GAN  to guide pruning, and~\citet{Wang2018HAQHA} structure compression as a reinforcement learning problem; both remove some of the burden from the user.

\section{Existing Techniques Fail for Complex Task}\label{sec-naiveGANCompression}

\begin{figure*}[!htb]
\begin{center}
    \centering
    \subfigtopskip=1pt
    \subfigcapskip=-2pt
    \subfigbottomskip=3pt
    \subfigure[][]{
    \label{Fig.1-dense}
    	\begin{minipage}[b]{0.48\linewidth}
    		\centering
    		\includegraphics[width = 0.89\linewidth]{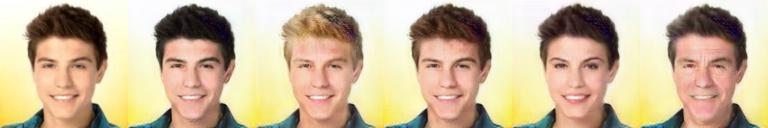}
    	\end{minipage}
    }
    \subfigure[][]{
    \label{Fig.1-ours}
    	\begin{minipage}[b]{0.48\linewidth}
    		\centering
    		\includegraphics[width = 0.89\linewidth]{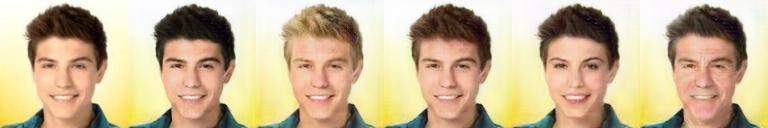}
    	\end{minipage}
    }
    \subfigure[][]{
    \label{Fig.1-small_dense}
    	\begin{minipage}[b]{0.48\linewidth}
    		\centering
    		\includegraphics[width = 0.89\linewidth]{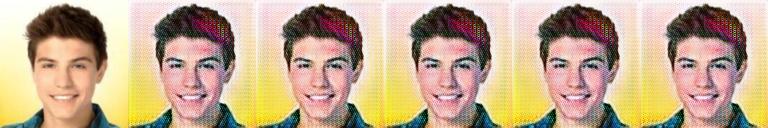}
    	\end{minipage}
    }
    \subfigure[][]{
    \label{Fig.1-oneShot}
    	\begin{minipage}[b]{0.48\linewidth}
    		\centering
    		\includegraphics[width = 0.89\linewidth]{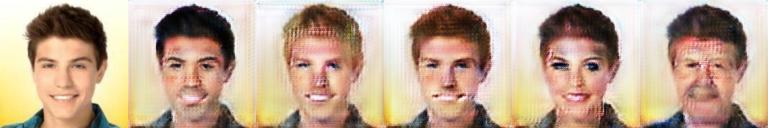}
    	\end{minipage}
    }
    \subfigure[][]{
    \label{Fig.1-AGP-FT}
    	\begin{minipage}[b]{0.48\linewidth}
    		\centering
    		\includegraphics[width = 0.89\linewidth]{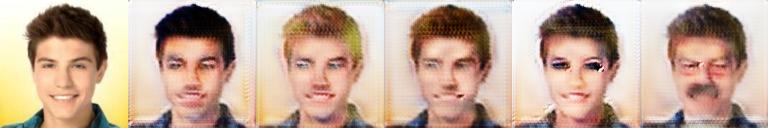}
    	\end{minipage}
    }
    \subfigure[][]{
    \label{Fig.1-AGP-S}
    	\begin{minipage}[b]{0.48\linewidth}
    		\centering
    		\includegraphics[width = 0.89\linewidth]{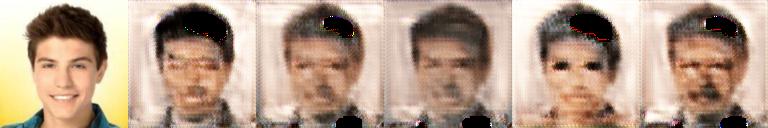}
    	\end{minipage}
    }
    \subfigure[][]{
    \label{Fig.1-oneShot-distill}
    	\begin{minipage}[b]{0.48\linewidth}
    		\centering
    		\includegraphics[width = 0.89\linewidth]{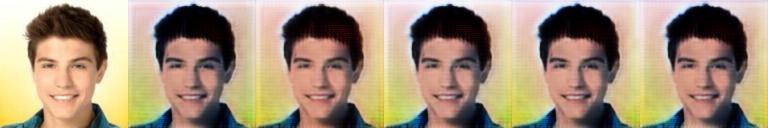}
    	\end{minipage}
    }
    \subfigure[][]{
    \label{Fig.1-AGP_distill}
    	\begin{minipage}[b]{0.48\linewidth}
    		\centering
    		\includegraphics[width = 0.89\linewidth]{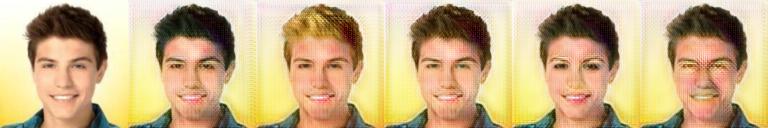}
    	\end{minipage}
    }
    \subfigure[][]{
    \label{Fig.1-AGP_distill_fixed}
    	\begin{minipage}[b]{0.48\linewidth}
    		\centering
    		\includegraphics[width = 0.89\linewidth]{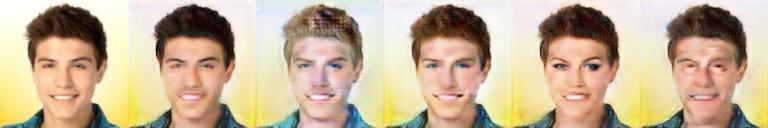}
    	\end{minipage}
    }
    \subfigure[][]{
    \label{Fig.1-Adversarial}
    	\begin{minipage}[b]{0.48\linewidth}
    		\centering
    		\includegraphics[width = 0.89\linewidth]{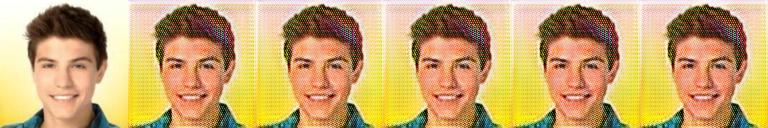}
    	\end{minipage}
    }
    \subfigure[][]{
    \label{Fig.1-Aguinaldo}
    	\begin{minipage}[b]{0.48\linewidth}
    		\centering
    		\includegraphics[width = 0.89\linewidth]{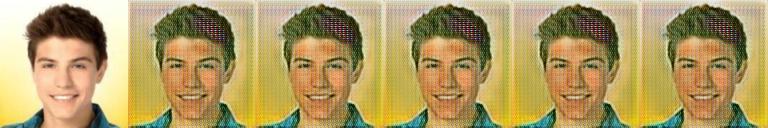}
    	\end{minipage}
    }
    \subfigure[][]{
    \label{Fig.1-LIT}
    	\begin{minipage}[b]{0.48\linewidth}
    		\centering
    		\includegraphics[width = 0.89\linewidth]{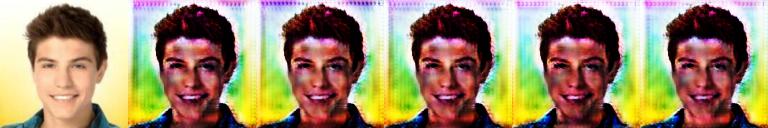}
    	\end{minipage}
    }
    \subfigure[][]{
    \label{Fig.1-EMQuant}
    	\begin{minipage}[b]{0.48\linewidth}
    		\centering
    		\includegraphics[width = 0.89\linewidth]{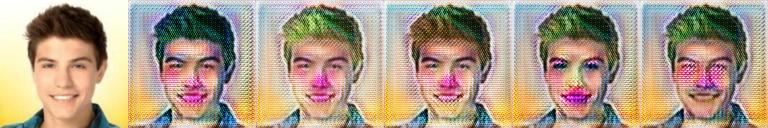}
    	\end{minipage}
    }
    \subfigure[][]{
    \label{Fig.1-pruneBoth}
    	\begin{minipage}[b]{0.48\linewidth}
    		\centering
    		\includegraphics[width = 0.89\linewidth]{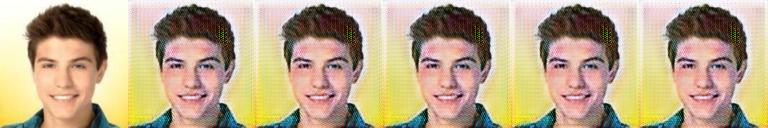}
    	\end{minipage}
    }
\end{center}
\caption{Various approaches to compress StarGAN with network pruning. Each group shows one input face translated with different methods of compressing the network: \textbf{a}. Uncompressed, \textbf{b}. Self-Supervised \textbf{(ours)}, \textbf{c}. Small and dense, \textbf{d}. One-shot pruning and fine-tuning, \textbf{e}. AGP as fine-tuning, \textbf{f}. AGP from scratch, \textbf{g}. One-shot pruning and distilling, \textbf{h}. AGP during distillation, \textbf{i}. AGP during distillation with fixed discriminator, \textbf{j}. Adversarial learning, \textbf{k}. Knowledge distillation, \textbf{l}. Distillation on output of intermediate layers, \textbf{m}. E-M pruning, and \textbf{n}. Prune both G and D models.}
\label{Fig.1}
\end{figure*}

Though there are two networks in a single GAN, the main workload at deployment is usually from the generative model, or generator. For example, in image synthesis and style transfer tasks, the final output images are created solely by the generator. The discriminative model (discriminator) is vital in training, but it is abandoned afterward for many tasks. So, when we try to apply state-of-the-art compression methods to GANs, we focus on the generator for efficient deployment. As we will see, the generative performance of the compressed generators is quite poor for the selected image-to-image translation task. We look at two broad categories of baseline approaches: standard pruning techniques that have been applied to other network architectures, and techniques that were devised to compress the generator of a GAN performing image synthesis. We compare the dense baseline [a] to our technique [b], as well as a small, dense network with the same number of parameters [c]. (Labels correspond to entries in Table~\ref{table:GANCompression}, the overview of all techniques, and Figure~\ref{Fig.1}, results of each technique).

\begin{table*}[!htb]
\caption{GAN compression comparison (network pruning)}
\label{table:GANCompression}
\centering
\resizebox{0.99\textwidth}{!}{%
\begin{tabular}{lcccccccccc}
\toprule
                                       &        \multicolumn{2}{c}{\textbf{Generator(s)}}        & \multicolumn{2}{c}{\textbf{Discriminator}}
                                       & \multicolumn{4}{c}{\textbf{Loss Terms}}                   & \multicolumn{2}{c}{\textbf{Results}} \\
{\bf Technique}                                & Compressed  & Init Scheme
                                                                          & Init Scheme         & Fixed
                                                                      & L-Gc    & L-Dc    & L-Go    & L-Do          & Qualitative         & FID Score\\
\midrule
(a) No Compression                     
											      & Dense        & Random
                                                                          & Dense,Random        & No
                                                                      & -      & -       & Yes     & Yes            & Good                &6.113 \\
(b) Self-Supervised \textbf{(ours)}    
												& Dense,Sparse & From Dense
                                                                          & Dense,Pretrained    & No
                                                                      & Yes    & Yes     & Yes     & Yes            & Good                &6.929 \\
(c) Small $\&$ Dense Network           
													& Dense        & Random
                                                                          & Dense,Random        & No
                                                                      & -      & -       & Yes     & Yes            & Mode collapse       & 72.821 \\
(d) One-shot Pruning $\&$ Fine-Tuning  
                         			                                  & Sparse       & From Dense
                                                                          & Dense,Pretrained    & No
                                                                      & Yes    & Yes     & -       & -              & Facial artifacts    & 24.404 \\
(e) Gradual Pruning $\&$ Fine-Tuning 
                                                                      & Sparse       & From Dense
                                                                          & Dense,Random        & No
                                                                      & Yes    & Yes     & -       & -              & Facial artifacts    & 35.677 \\
(f) Gradual Pruning during Training    
                                                                      & Sparse       & Random
                                                                          & Dense,Random		& No
                                                                      & Yes    & Yes     & -       & -              & No faces            & 84.941 \\
(g) One-shot Pruning $\&$ Distillation 
                                                                      & Dense,Sparse & From Dense
                                                                          & -                   & -
                                                                      & Yes    & -       & Yes     & -              & Mode collapse       & 45.461 \\
(h) (d) $\&$ Distillation              
                                                                      & Dense,Sparse & From Dense
                                                                          & Dense,Pretrained    & No
                                                                      & Yes    & Yes     & Yes     & -              & Color artifacts     & 38.985 \\
(i) (g) $\&$ Fix Original Loss         
                                                                      & Dense,Sparse & From Dense
                                                                          & Dense,Pretrained    & Yes
                                                                      & Yes    & Yes     & -       & -              & Facial artifacts    & 15.182 \\
(j) Adversarial Learning               
                                                                      & Dense,Sparse & Random
                                                                          & Dense,Random        & No
                                                                      & Yes    & Yes     & Yes     & Yes            & Mode collapse       & 92.721 \\
(k) Knowledge Distillation             
                                                                      & Dense,Sparse & From Dense
                                                                          & Dense,Random        & No
                                                                      & Yes    & -       & Yes     & Yes            & Mode collapse       & 103.094 \\
(l) Distill Intermediate (LIT)               
                                                                      & Dense,Sparse & From Dense
                                                                          & Dense,Pretrained    & Yes
                                                                      & -      & -       & -       & -              & Mode collapse       & 61.150 \\
(m) E-M Pruning                   
													& Dense,Sparse & From Dense
                                                                          & Sparse,Pretrained   & No
                                                                      & Yes    & Yes     & Yes     & -              & Color artifacts     & 159.767 \\
(n) G $\&$ D Both Pruning              
													& Dense,Sparse & From Dense
                                                                          & Sparse,Pretrained   & No
                                                                      & Yes    & Yes     & Yes     & -              & Mode collapse       & 46.453 \\
\bottomrule
\end{tabular}%
}
\end{table*}

\textbf{Standard Pruning Techniques}.
To motivate GAN-specific compression methods, we try variations of two state-of-the-art pruning methods: manually pruning and fine tuning~\citep{han2015learning} a trained dense model [d], and AGP~\citep{zhu2017prune} from scratch [e] and during fine-tuning [f]. We also include distillation~\citep{hinton2015distilling} to improve the performance of the pruned network with manual pruning [g] and AGP fine-tuning [h]. Distillation is typically optional for other network types, since it is possible to get decent accuracy with moderate pruning in isolation. For very aggressive\- compression or challenging tasks, distillation aims to extract knowledge for the compressed (student) network from original (teacher) network's behavior. We also fix the discriminator of [g] to see if the discriminator was being weakened by the compressed generator [i].

\textbf{Targeted GAN Compression}.
There has been some work in compressing GANs with methods other than pruning, and only one technique applied to an image-to-image translation task. For this category, we decompose each instance of prior work into two areas: the method of compression (e.g. quantization, layer removal, etc.) and the modifications required to make the compression succeed (e.g. distillation, novel training schemes, etc.). For comparisons to these techniques, we apply the modifications presented in prior research to the particular method of compression on which we focus, network pruning. We first examine two approaches similar to ours. Adversarial training~\citep{wang2018adversarial} [j] posits that during distillation of a classification network, the student network can be thought of as a generative model attempting to produce features similar to that of the teacher model. So, a discriminator was trained alongside the student network, trying to distinguish between the student and the teacher. One could apply this technique to compress the generator of a GAN, but we find that its key shortcoming is that it trains a discriminator from scratch. Similarly, distillation has been used to compress GANs in~\citet{aguinaldo2019compressing} [k], but again, the ``teacher" discriminator was not used when teaching the ``student" generator.

Learned Intermediate Representation Training (LIT)~\citep{koratana2019LIT} [l] compresses StarGAN by a factor of $1.8\times$ by training a shallower network. Crucially, LIT does not use the pre-trained discriminator in any loss function. Quantized GANs (QGAN)~\citep{Wang2019QGAN} [m] use a training process based on Expectation-Maximization to achieve impressive compression results on small generative tasks with output images of 32x32 or 64x64 pixels. ~\citet{Liu2018PBGen} find that maintaining a balance between discriminator and generator is key: their approach is to selectively binarize parts of both networks in the training process on the Celeb-A generative task, up to 64x64 pixels. So, we try pruning both networks during the training process [n].

\textbf{Experiments}.
For these experiments, we use StarGAN\footnote{StarGAN:
\url{https://github.com/yunjey/StarGAN}.}~\citep{choi2018stargan} trained with the Distiller~\citep{neta_zmora_2018_1297430} library for the pruning. StarGAN extends the image-to-image translation capability from two domains to multiple domains within a single unified model. It uses the CelebFaces Attributes (CelebA)~\citep{liu2015deep} as the dataset. CelebA contains 202,599 images of celebrities' faces, each annotated with 40 binary attributes. As in the original work, we crop the initial images from size $178\times218$ to $178\times178$, then resize them to $128\times128$ and randomly select 2,000 images as the test dataset and use remaining images for training. The aim of StarGAN is facial attribute translation: given some image of a face, it generates new images with five domain attributes changed: 3 different hair colors (black, blond, brown), different gender (male/female), and different age (young/old). Our target sparsity is 50\% for each approach.

We stress that we attempted to find good hyperparameters when using the existing techniques, but standard approaches like reducing the learning rate for fine-tuning~\citep{han2015learning}, etc., were not helpful. Further, the target sparsity, 50\%, is not overly aggressive, and we do not impose any structure; other tasks readily achieve 80\%-90\% fine-grained sparsity with minimal accuracy impact.

The results of these trials are shown in Figure~\ref{Fig.1}. Subjectively, it is easy to see that the existing approaches (\ref{Fig.1-small_dense} through \ref{Fig.1-pruneBoth}) produce inferior results to the original, dense generator. Translated facial images from pruning $\&$ na\"ive fine-tuning (\ref{Fig.1-oneShot} and~\ref{Fig.1-AGP-FT}) do give unique results for each latent variable, but the images are hardly recognizable as faces. These fine-tuning procedures, along with AGP from scratch (\ref{Fig.1-AGP-S}) and distillation from intermediate representations (\ref{Fig.1-LIT}), simply did not converge. One-shot pruning and traditional distillation (\ref{Fig.1-oneShot-distill}), adversarial learning (\ref{Fig.1-Adversarial}), knowledge distillation (\ref{Fig.1-Aguinaldo}), training a ``smaller, dense" half-sized network from scratch (\ref{Fig.1-small_dense}) and pruning both generator and discriminator (\ref{Fig.1-pruneBoth}) keep facial features intact, but the image-to-image translation effects are lost to mode collapse (see below). There are obvious mosaic textures and color distortion on the translated images from fine-tuning $\&$ distillation (\ref{Fig.1-AGP_distill}), without fine-tuning the original loss (\ref{Fig.1-AGP_distill_fixed}), and from the pruned model based on the Expectation-Maximization (E-M) algorithm (\ref{Fig.1-EMQuant}). However, the translated facial images from a generator compressed with our proposed self-supervised GAN compression method (\ref{Fig.1-ours}) are more natural, nearly indistinguishable from the dense baseline (\ref{Fig.1-dense}), matching the quantitative Frechet Inception Distance (FID) scores~\citep{heusel2017gans} in Table \ref{table:GANCompression}. While past approaches have worked to prune some networks on other tasks (DCGAN generating MNIST digits, see the supplementary material), we show that they do not succeed on larger image-to-image translation tasks, while our approach works on both. Similarly, though LIT~\citep{koratana2019LIT} [l] was able to achieve a compression rate of $1.8\times$ on this task by training a shallower network, it does not see the same success at network pruning with a higher rate, as modest as 50\% sparsity is.

\textbf{Discussion}.
It is tempting to think that the loss curves of the experiment for each technique can tell us if the result is good or not. We found that for many of these experiments, the loss curves correctly predicted that the final result would be poor. However, the curves for [h] and [m] look very good - the compressed generator and discriminator losses converge at 0, just as they did for baseline training. It is clear from the results of querying the generative models (Figures~\ref{Fig.1-AGP_distill} and~\ref{Fig.1-EMQuant}), though, that this promising convergence is a false positive. In contrast, the curves for our technique predict good performance, and, as we prune more aggressively in Section~\ref{sec-pruningRatioGranularity}, higher loss values correlate well with worsening FID scores. (Loss curves are provided in the \emph{\textbf{Appendix}}.)

As pruning and distillation are very effective when compressing models for image classification tasks, why do they fail to compress this generative model? We share three potential reasons:
\begin{enumerate}
\vspace{-1em}
\setlength{\itemsep}{0pt}
\setlength{\parsep}{0pt}
\setlength{\parskip}{0pt}
\item Standard pruning techniques need explicit evaluation metrics; softmax easily reflects the probability distribution and classification accuracy. GANs are typically evaluated subjectively, though some imperfect quantitative metrics have been devised.
\item GAN training is relatively unstable~\citep{arjovsky2017wasserstein,Liu2018PBGen} and sensitive to hyperparameters. The generator and discriminator must be well-matched, and pruning can disrupt this fine balance.
\item The energy of the input and output of a GAN is roughly constant, but other tasks, such as classification, produce an output (1-hot label vector) with much less entropy than the input (three-channel color image of thousands of pixels).
\vspace{-1em}
\end{enumerate}

Elaborating on this last point, there is more tolerance in the reduced-information space for the compressed classification model to give the proper output. That is, even if the probability distribution inferred by the original and compressed classification models are not exactly the same, the classified labels \emph{can} be the same. On the other hand, tasks like style-transfer and dataset synthesis have no obvious energy reduction. We need to keep entropy as high as possible~\citep{kumar2019maximum} during the compression process to avoid mode collapse -- generating the same output for different inputs or tasks. Attempting to train a new discriminator to make the compressed generator behave more like the original generator~\citep{wang2018adversarial} suffers from this issue -- the new discriminator quickly falls into a low-entropy solution and cannot escape. Not only does this preclude its use on generative tasks, but it means that the compressed network for any task must also be trained from scratch during the distillation process, or the discriminator will never be able to learn.


\section{Self-Supervised Generator Compression}\label{sec-self-supervisedGeneratorCompression}
\begin{figure*}[htb]
  \centering
  \includegraphics[width=0.95\textwidth]{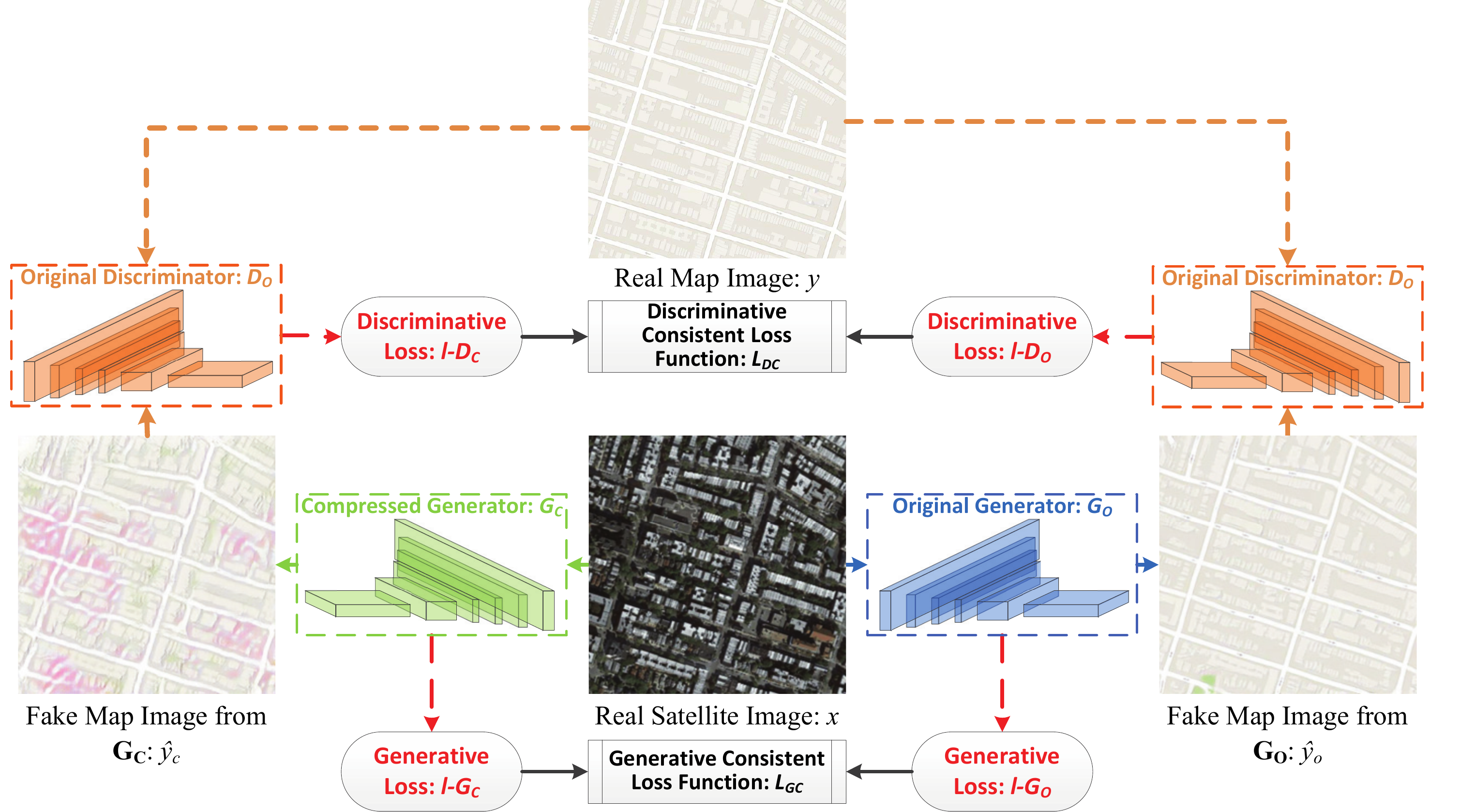}\\
  \caption{Workflow chart of GAN compression process.}
  \label{Fig.Workflow}
\end{figure*}

We seek to solve each of the problems highlighted above. Let us restate the general formulation of GAN training: the purpose of the generative model is to generate new samples which are very similar to the real samples, but the purpose of the \emph{discriminative} model is to distinguish between real samples and those synthesized by the generator. A fully-trained discriminator is good at spotting differences, but a well-trained generator will cause it to believe that the a generated sample is both real and generated with a probability of 0.5. Our main insight follows:

By using this powerful discriminator that is already well-trained on the target data set, we can allow it to stand in as a quantitative subjective judge (point 1, above) -- if the discriminator can't tell the difference between real data samples and those produced by the compressed generator, then the compressed generator is of the same quality as the uncompressed generator. A human no longer needs to inspect the results to judge the quality of the compressed generator. This also addresses our second point: by starting with a trained discriminator, we know it is well-matched to the generator and will not be overpowered. Since it is so capable (there is no need to prune it to), it also helps to avoid mode collapse. As distillation progresses, it can adapt to and induce fine changes in the compressed generator, which is initialized from the uncompressed generator.
Since the original discriminator is used as a proxy for a human's subjective evaluation, we refer to this as ``self-supervised" compression. We illustrate the workflow in Figure~\ref{Fig.Workflow}, using a GAN charged with generating a map image from a satellite image in a domain translation task.

In the right part of Figure~\ref{Fig.Workflow}, the real satellite image (\emph{x}) goes through the original generative model (\textbf{\emph{G}$_{O}$}) to produce a fake map image (\emph{$\hat{y}$}$_{o}$). The corresponding generative loss value is \emph{l-}\textbf{\emph{G}$_{O}$}. Accordingly, in the left part of Figure~\ref{Fig.Workflow}, the real satellite image (\emph{x}) goes through the compressed generative model (\textbf{\emph{G}$_{C}$}) to produce a fake map image (\emph{$\hat{y}$}$_{c}$). The corresponding generative loss value is \emph{l-}\textbf{\emph{G}$_{C}$}. This is the inference process of the original and compressed generators, expressed as follows:
\begin{equation}
\textbf{\emph{$\hat{y}$}$_{o}$} = \textbf{\emph{G$_{O}$}}(x), \qquad \textbf{\emph{$\hat{y}$}$_{c}$} = \textbf{\emph{G$_{C}$}}(x)
\end{equation}
The overall generative difference is measured between the two corresponding generative losses\footnote{In different GANs, the generative loss may consist of several sub-items. For example, StarGAN combines adversarial loss, domain classification loss and reconstruction loss into overall generative loss.}. We use a generative consistent loss function (\textbf{\emph{L$_{GC}$}}) in the bottom of Figure~\ref{Fig.Workflow} to represent this process.
\begin{equation}
\textbf{\emph{L$_{GC}$}}(\emph{l-}\textbf{\emph{G}$_{O}$},\emph{l-}\textbf{\emph{G}$_{C}$}) \rightarrow 0
\end{equation}
Since the GAN training process aims to reduce the differences between real and generated samples, we stick to this principle in the compression process. In the upper right of Figure~\ref{Fig.Workflow}, real map image (\emph{y}) and fake map image (\emph{$\hat{y}$}$_{o}$) go through the original discriminative model \textbf{\emph{D}$_{O}$}. \textbf{\emph{D}$_{O}$} tries to ensure that the distribution of \emph{$\hat{y}$}$_{o}$ is indistinguishable from \emph{y} using an adversarial loss. The corresponding discriminative loss value is \emph{l-}\textbf{\emph{D}$_{O}$}. In the upper left of Figure~\ref{Fig.Workflow}, real map image (\emph{y}) and fake map image (\emph{$\hat{y}$}$_{c}$) also go through the original discriminative model \textbf{\emph{D}$_{O}$}. In this way, we use the original discriminative model as a ``self-supervisor". The corresponding discriminative loss value is \emph{l-}\textbf{\emph{D}$_{C}$}.
\begin{equation}
\emph{l-}\textbf{\emph{D}$_{O}$}= \textbf{\emph{D$_{O}$}}(\textbf{\emph{$y$}}, \textbf{\emph{$\hat{y}$}$_{o}$}), \qquad \emph{l-}\textbf{\emph{D}$_{C}$}= \textbf{\emph{D$_{O}$}}(\textbf{\emph{$y$}}, \textbf{\emph{$\hat{y}$}$_{c}$})
\end{equation}
So the discriminative difference is measured between two corresponding discriminative losses. We use the discriminative consistent loss function \textbf{\emph{L$_{DC}$}} in the top of Figure~\ref{Fig.Workflow} to represent this process.
\begin{equation}
\textbf{\emph{L$_{DC}$}}(\emph{l-}\textbf{\emph{D}$_{O}$},\emph{l-}\textbf{\emph{D}$_{C}$}) \rightarrow 0
\end{equation}
The generative and discriminative consistent loss functions (\textbf{\emph{L$_{GC}$}} and \textbf{\emph{L$_{DC}$}}) use the weighted normalized Euclidean distance. Taking the \textbf{StarGAN} task as the example (other tasks may use different losses):
\begin{equation}
\begin{aligned}
\textbf{\emph{L$_{GC}$}}(\emph{l-}\textbf{\emph{G}$_{O}$},\emph{l-}\textbf{\emph{G}$_{C}$}) =
|\emph{l-}\textbf{\emph{Gen}$_{O}$} - \emph{l-}\textbf{\emph{Gen}$_{C}$}|/|\emph{l-}\textbf{\emph{Gen}$_{O}$}| + \\ \alpha|\emph{l-}\textbf{\emph{Cla}$_{O}$} - \emph{l-}\textbf{\emph{Cla}$_{C}$}|/|\emph{l-}\textbf{\emph{Cla}$_{O}$}| + \\
\beta|\emph{l-}\textbf{\emph{Rec}$_{O}$} - \emph{l-}\textbf{\emph{Rec}$_{C}$}|/|\emph{l-}\textbf{\emph{Rec}$_{O}$}|
\end{aligned}
\end{equation}
where \emph{l-}\textbf{\emph{Gen}} is the generation loss term, \emph{l-}\textbf{\emph{Cla}} is the classification loss term, and \emph{l-}\textbf{\emph{Rec}} is the reconstruction loss term. $\alpha$ and $\beta$ are the weight ratios among three loss types. (We use the same values of $\alpha$ and $\beta$ used in the original StarGAN baseline.)
\begin{equation}
\begin{aligned}
\textbf{\emph{L$_{DC}$}}(\emph{l-}\textbf{\emph{D}$_{O}$},\emph{l-}\textbf{\emph{D}$_{C}$}) =
|\emph{l-}\textbf{\emph{Dis}$_{O}$} - \emph{l-}\textbf{\emph{Dis}$_{C}$}|/|\emph{l-}\textbf{\emph{Dis}$_{O}$}| + \\
\delta|\emph{l-}\textbf{\emph{GP}$_{O}$} - \emph{l-}\textbf{\emph{GP}$_{C}$}|/|\emph{l-}\textbf{\emph{GP}$_{O}$}|
\end{aligned}
\end{equation}
where \emph{l-}\textbf{\emph{Dis}} is the discriminative loss item, \emph{l-}\textbf{\emph{GP}} is the gradient penalty loss item, and $\delta$ is a weighting factor (again, we use the same value as the baseline).

The overall loss function of GAN compression consists of generative and discriminative differences:
\begin{equation}
\emph{L}_{Overall} = \textbf{\emph{L$_{GC}$}}(\emph{l-}\textbf{\emph{G}$_{O}$},\emph{l-}\textbf{\emph{G}$_{C}$}) + \lambda\textbf{\emph{L$_{DC}$}}(\emph{l-}\textbf{\emph{D}$_{O}$},\emph{l-}\textbf{\emph{D}$_{C}$}),
\end{equation}
where $\lambda$ is the parameter to adjust the percentages between generative and discriminative losses.

We showed promising results with this method above in the context of prior methods. In the following experiments, we investigate how well the method applies to other networks and tasks (Section~\ref{sec-generalizationExperiments}) and how well the method works on different sparsity ratios and pruning granularities (Section~\ref{sec-pruningRatioGranularity}).

\section{Application to New Tasks and Networks}\label{sec-generalizationExperiments}
For the experiments in this section, we choose to prune individual weights in the generator. The final sparsity rate is 50\% for all convolution and deconvolution layers in the generator (more aggressive sparsities are discussed in Section~\ref{sec-pruningRatioGranularity}). Following AGP~\citep{zhu2017prune}, we gradually increase the sparsity from 5\% at the beginning to our target of 50\% halfway through the self-supervised training process, and we set the loss adjustment parameter $\lambda$ to 0.5 in all experiments. We use PyTorch~\citep{paszke2017automatic}, implement the pruning and training schedules with Distiller~\citep{neta_zmora_2018_1297430}, and train and generate results with a V100 GPU~\citep{nvidiav100} using FP32 to match public baselines.
In all experiments, the data sets, data preparation, and baseline training all follow from the public repositories - details are summarized in Table~\ref{table:TaskNetworkOverview}. We start by assuming an extra 10\% of the original number of epochs will be required; in some cases, we reduced the overhead to only 1\% while maintaining subjective quality. We include representative results for each task, but a more comprehensive collection of outputs for each experiment is included in the \emph{\textbf{Appendix}}.

\begin{table*}[htb]
\caption{Tasks and networks overview}
\label{table:TaskNetworkOverview}
\centering
\resizebox{0.99\textwidth}{!}{%
\begin{tabular}{llllccccc}
\toprule
\multirow{2}{*}{Task}  & \multirow{2}{*}{Network}  & \multirow{2}{*}{Dataset} & \multirow{2}{*}{Resolution} & \multicolumn{5}{c}{FID Scores when Pruned to} \\
\cmidrule(lr){5-9}
                                     &                                            &                                            &                                                & 0\% (dense) &     25\%       &    50\%      &     75\%      &     90\% \\
\midrule
Image Synthesis         & DCGAN            & MNIST                           & 64x64               & 50.391 & 50.128 & 50.634 & 50.805 & 51.356 \\
Domain Translation      & Pix2Pix          & Sat $\rightarrow$ Map           & 256x256             & 17.636 & 17.897 & 17.990 & 20.235 & 24.892 \\
Domain Translation      & Pix2Pix          & Sat $\leftarrow$ Map            & 256x256             & 30.826 & 30.628 & 30.720 & 34.051 & 38.936 \\
Style Transfer          & CycleGAN         & Monet $\rightarrow$ Photo       & 256x256             & 63.152 & 63.410 & 63.662 & 66.394 & 70.933 \\
Style Transfer          & CycleGAN         & Monet $\leftarrow$ Photo        & 256x256             & 31.987 & 32.102 & 32.346 & 33.913 & 41.409 \\
Image-Image Translation & CycleGAN         & Zebra $\rightarrow$ Horse       & 256x256             & 60.930 & 61.005 & 61.102 & 65.898 & 68.450 \\
Image-Image Translation & CycleGAN         & Zebra $\leftarrow$ Horse        & 256x256             & 52.862 & 52.631 & 52.688 & 58.356 & 63.274 \\
Image-Image Translation & StarGAN          & CelebA                          & 128x128             & 6.113  & 6.307  & 6.929  & 6.714  & 7.144  \\
Super Resolution        & SRGAN            & DIV2K                           & $\geq$ 512x512      & 14.653 & 15.236 & 16.609 & 17.548 & 18.376 \\
\bottomrule
\end{tabular}%
}
\end{table*}

\begin{figure}[htb]
\begin{center}
    \centering
	\begin{minipage}[b]{0.32\linewidth}
		\centering
		\includegraphics[width = 0.95\linewidth]{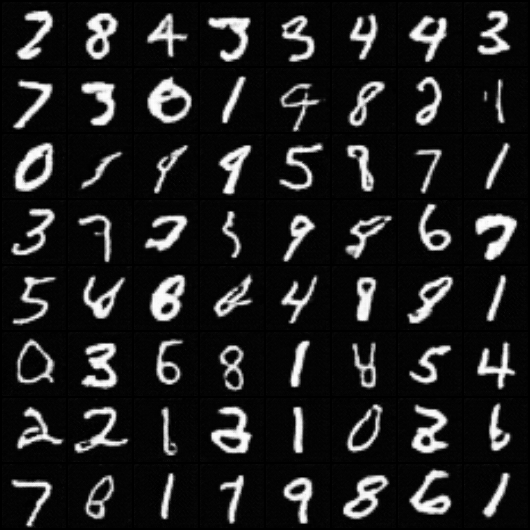}
    \textbf{\emph{\texttt{\fontsize{6.5pt}{\baselineskip}\selectfont FID: 32.7786}}}
    \end{minipage}
	\begin{minipage}[b]{0.32\linewidth}
		\centering
		\includegraphics[width = 0.95\linewidth]{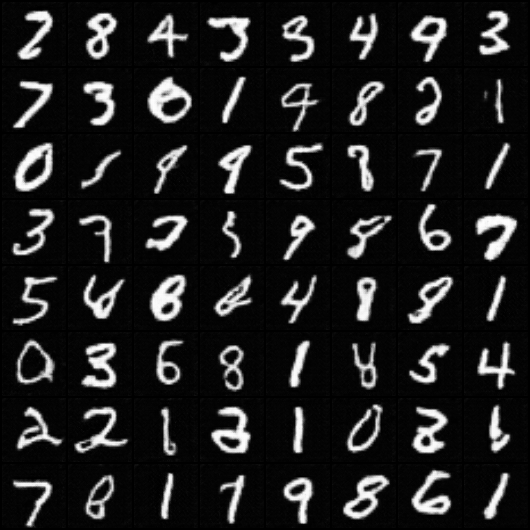}	
    \textbf{\emph{\texttt{\fontsize{6.5pt}{\baselineskip}\selectfont 33.3191}}}
    \end{minipage}
    \begin{minipage}[b]{0.32\linewidth}
		\centering
		\includegraphics[width = 0.95\linewidth]{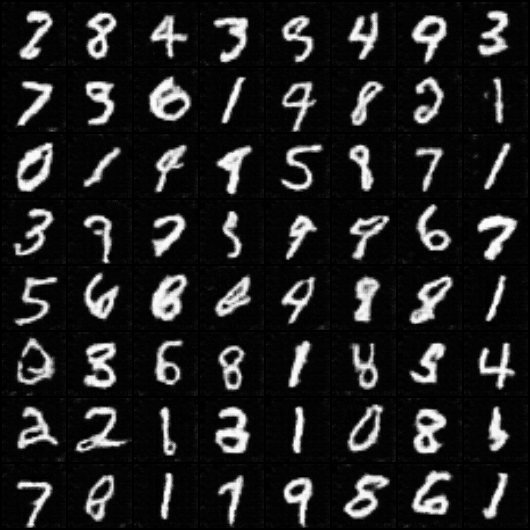}	
    \textbf{\emph{\texttt{\fontsize{6.5pt}{\baselineskip}\selectfont 37.1807}}}
    \end{minipage}
\end{center}
	\caption{Image synthesis on MNIST dataset with DCGAN. Columns 1-2: Handwritten numbers generated by the original generator and pruned generator of 50\%, 75\% fine-grained sparsity.}
	\label{Fig.MNIST_DCGAN}
\end{figure}

\textbf{Image Synthesis}. We apply the proposed compression method to DCGAN~\citep{radford2015unsupervised}\footnote{DCGAN baseline:
\url{https://github.com/pytorch/examples/tree/master/dcgan}.}, a network that learns to synthesize novel images from some distribution. We task DCGAN with generating images that could belong to the MNIST data set, with results shown in Figure~\ref{Fig.MNIST_DCGAN}.

\begin{figure}[t]
\begin{center}
    \centering
    \begin{minipage}[b]{0.24\linewidth}
		\centering
		\textbf{\texttt{\fontsize{5.5pt}{\baselineskip}\selectfont Input}}
	\end{minipage}
    \begin{minipage}[b]{0.24\linewidth}
		\centering
		\textbf{\texttt{\fontsize{5.5pt}{\baselineskip}\selectfont Dense}}
	\end{minipage}
	\begin{minipage}[b]{0.24\linewidth}
		\centering
		\textbf{\texttt{\fontsize{5.5pt}{\baselineskip}\selectfont 50\% Sparse}}
	\end{minipage}
	\begin{minipage}[b]{0.24\linewidth}
		\centering
		\textbf{\texttt{\fontsize{5.5pt}{\baselineskip}\selectfont Difference (10x)}}
	\end{minipage}

    	\begin{minipage}[b]{0.24\linewidth}
    		\centering
    		\includegraphics[width = 0.99\linewidth]{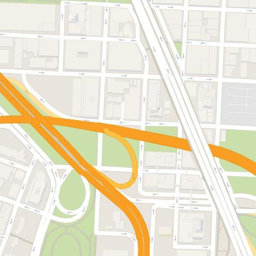}
    	\vskip -0.05in
        \textbf{\emph{\texttt{\fontsize{6.5pt}{\baselineskip}\selectfont FID:}}}
        \end{minipage}
    	\begin{minipage}[b]{0.24\linewidth}
    		\centering
    		\includegraphics[width = 0.99\linewidth]{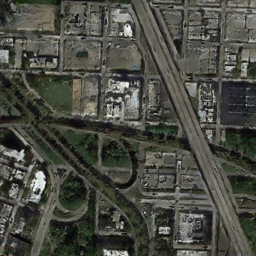}
        \vskip -0.05in
        \textbf{\emph{\texttt{\fontsize{6.5pt}{\baselineskip}\selectfont 35.627}}}
        \end{minipage}
    	\begin{minipage}[b]{0.24\linewidth}
    		\centering
    		\includegraphics[width = 0.99\linewidth]{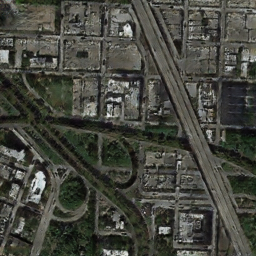}
    	\vskip -0.05in
        \textbf{\emph{\texttt{\fontsize{6.5pt}{\baselineskip}\selectfont 33.366}}}
        \end{minipage}
    	\begin{minipage}[b]{0.24\linewidth}
    		\centering
    		\includegraphics[width = 0.99\linewidth]{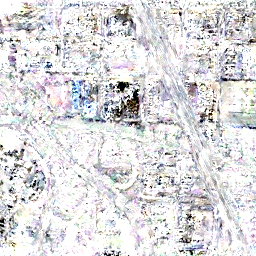}    	
        \vskip -0.05in		
        \textbf{\emph{\texttt{\fontsize{6.5pt}{\baselineskip}\selectfont -}}}
        \end{minipage}

    	\begin{minipage}[b]{0.24\linewidth}
    		\centering
    		\includegraphics[width = 0.99\linewidth]{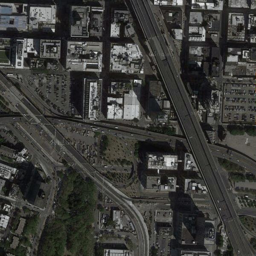}	
    	\vskip -0.05in
        \textbf{\emph{\texttt{\fontsize{6.5pt}{\baselineskip}\selectfont FID:}}}
        \end{minipage}
    	\begin{minipage}[b]{0.24\linewidth}
    		\centering
    		\includegraphics[width = 0.99\linewidth]{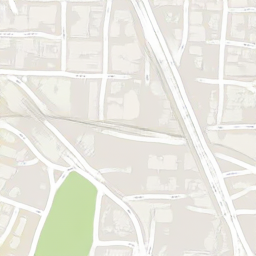}	
    	\vskip -0.05in
        \textbf{\emph{\texttt{\fontsize{6.5pt}{\baselineskip}\selectfont 17.097}}}
        \end{minipage}
    	\begin{minipage}[b]{0.24\linewidth}
    		\centering
    		\includegraphics[width = 0.99\linewidth]{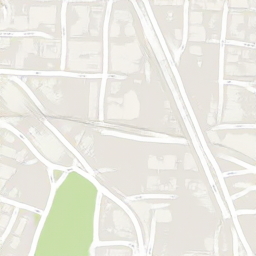}	
    	\vskip -0.05in
        \textbf{\emph{\texttt{\fontsize{6.5pt}{\baselineskip}\selectfont 17.945}}}
        \end{minipage}
    	\begin{minipage}[b]{0.24\linewidth}
    		\centering
    		\includegraphics[width = 0.99\linewidth]{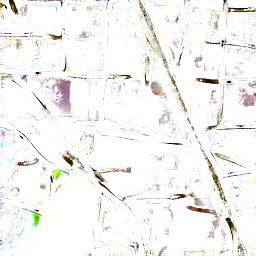}    	
		\vskip -0.05in
        \textbf{\emph{\texttt{\fontsize{6.5pt}{\baselineskip}\selectfont -}}}
        \end{minipage}
\end{center}
	\caption{Representative results for domain translation: pix2pix. Row 1: map to satellite task, Row 2: satellite to map task.}
	\label{Fig-cyclegan}
\end{figure}

\textbf{Domain Translation}. We apply the proposed compression method to pix2pix~\citep{isola2017image}\footnote{Pix2pix, CycleGAN baseline:
\url{https://github.com/junyanz/pytorch-CycleGAN-and-pix2pix}.}, an approach to learn the mapping between paired training examples by applying conditional adversarial networks.
In our experiment, the task is synthesizing fake satellite images from label maps and vice-versa. Representative results of this bidirectional task are shown in Figure~\ref{Fig-cyclegan}.

\begin{figure}[htb]
\begin{center}
    \centering
       \begin{minipage}[b]{0.24\linewidth}
    		\centering
    		\textbf{\texttt{\fontsize{5.5pt}{\baselineskip}\selectfont Input}}
    	\end{minipage}
    	\begin{minipage}[b]{0.24\linewidth}
    		\centering
    		\textbf{\texttt{\fontsize{5.5pt}{\baselineskip}\selectfont Dense}}
    	\end{minipage}
    	\begin{minipage}[b]{0.24\linewidth}
    		\centering
    		\textbf{\texttt{\fontsize{5.5pt}{\baselineskip}\selectfont 50\% Sparse}}
    	\end{minipage}
    	\begin{minipage}[b]{0.24\linewidth}
    		\centering
    		\textbf{\texttt{\fontsize{5.5pt}{\baselineskip}\selectfont Difference (10x)}}
    	\end{minipage}

    	\begin{minipage}[b]{0.24\linewidth}
    		\centering
    		\includegraphics[width = 0.99\linewidth]{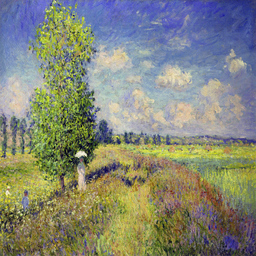}
    	\vskip -0.05in
        \textbf{\emph{\texttt{\fontsize{6.5pt}{\baselineskip}\selectfont FID:}}}
        \end{minipage}
    	\begin{minipage}[b]{0.24\linewidth}
    		\centering
    		\includegraphics[width = 0.99\linewidth]{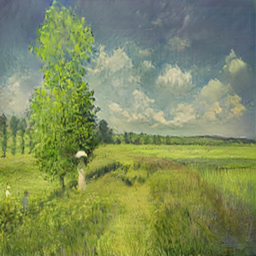}
        \vskip -0.05in
        \textbf{\emph{\texttt{\fontsize{6.5pt}{\baselineskip}\selectfont 59.381}}}
        \end{minipage}
    	\begin{minipage}[b]{0.24\linewidth}
    		\centering
    		\includegraphics[width = 0.99\linewidth]{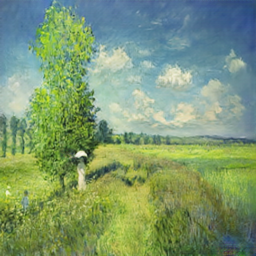}
    	\vskip -0.05in
        \textbf{\emph{\texttt{\fontsize{6.5pt}{\baselineskip}\selectfont 60.546}}}
        \end{minipage}
    	\begin{minipage}[b]{0.24\linewidth}
    		\centering
    		\includegraphics[width = 0.99\linewidth]{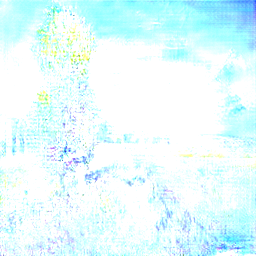}
		\vskip -0.05in
        \textbf{\emph{\texttt{\fontsize{6.5pt}{\baselineskip}\selectfont -}}}
        \end{minipage}

    	\begin{minipage}[b]{0.24\linewidth}
    		\centering
    		\includegraphics[width = 0.99\linewidth]{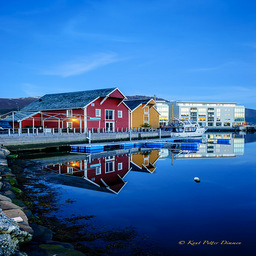}	
    	\vskip -0.05in
        \textbf{\emph{\texttt{\fontsize{6.5pt}{\baselineskip}\selectfont FID:}}}
        \end{minipage}
    	\begin{minipage}[b]{0.24\linewidth}
    		\centering
    		\includegraphics[width = 0.99\linewidth]{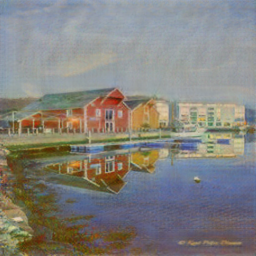}	
    	\vskip -0.05in
        \textbf{\emph{\texttt{\fontsize{6.5pt}{\baselineskip}\selectfont 35.781}}}
        \end{minipage}
    	\begin{minipage}[b]{0.24\linewidth}
    		\centering
    		\includegraphics[width = 0.99\linewidth]{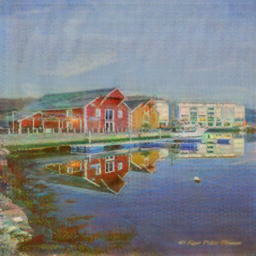}	
    	\vskip -0.05in
        \textbf{\emph{\texttt{\fontsize{6.5pt}{\baselineskip}\selectfont 34.263}}}
        \end{minipage}
    	\begin{minipage}[b]{0.24\linewidth}
    		\centering
    		\includegraphics[width = 0.99\linewidth]{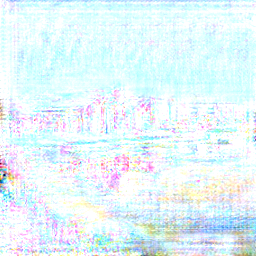}
		\vskip -0.05in
        \textbf{\emph{\texttt{\fontsize{6.5pt}{\baselineskip}\selectfont -}}}
        \end{minipage}
\end{center}
	\caption{Representative results for style transfer: CycleGAN. Row 1: Monet painting to photo, Row 2: photo to Monet painting.}
	\label{Fig.6}
\end{figure}

\textbf{Style Transfer}.
We apply the proposed compression method to CycleGAN~\citep{zhu2017unpaired}, used to exchange the style of images from a source domain to a target domain in the absence of paired training examples. In our experiment, the task is to transfer the style of real photos with that of the Monet's paintings.
Representative results of this bidirectional task are shown in Figure \ref{Fig.6}: photographs are given the style of Monet's paintings and vice-versa.

\begin{figure}[!htb]
\begin{center}
    \centering
        \begin{minipage}[b]{0.24\linewidth}
    		\centering
    		\textbf{\texttt{\fontsize{5.5pt}{\baselineskip}\selectfont Input}}
    	\end{minipage}
    	\begin{minipage}[b]{0.24\linewidth}
    		\centering
    		\textbf{\texttt{\fontsize{5.5pt}{\baselineskip}\selectfont Dense}}
    	\end{minipage}
    	\begin{minipage}[b]{0.24\linewidth}
    		\centering
    		\textbf{\texttt{\fontsize{5.5pt}{\baselineskip}\selectfont 50\% Sparse}}
    	\end{minipage}
    	\begin{minipage}[b]{0.24\linewidth}
    		\centering
    		\textbf{\texttt{\fontsize{5.5pt}{\baselineskip}\selectfont Difference (10x)}}
    	\end{minipage}

    	\begin{minipage}[b]{0.24\linewidth}
    		\centering
    		\includegraphics[width = 0.99\linewidth]{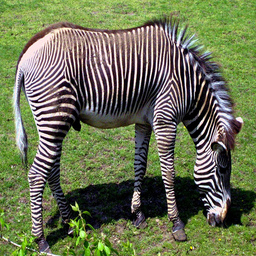}
    	\vskip -0.05in
        \textbf{\emph{\texttt{\fontsize{6.5pt}{\baselineskip}\selectfont FID:}}}
        \end{minipage}
    	\begin{minipage}[b]{0.24\linewidth}
    		\centering
    		\includegraphics[width = 0.99\linewidth]{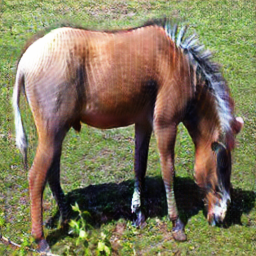}
    	\vskip -0.05in
        \textbf{\emph{\texttt{\fontsize{6.5pt}{\baselineskip}\selectfont 47.929}}}
        \end{minipage}
    	\begin{minipage}[b]{0.24\linewidth}
    		\centering
    		\includegraphics[width = 0.99\linewidth]{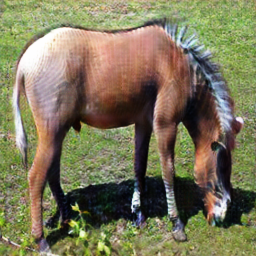}
    	\vskip -0.05in
        \textbf{\emph{\texttt{\fontsize{6.5pt}{\baselineskip}\selectfont 48.112}}}
        \end{minipage}
    	\begin{minipage}[b]{0.24\linewidth}
    		\centering
            \setlength{\fboxrule}{0.4pt}        \setlength{\fboxsep}{0cm}
    		\fbox{\includegraphics[width = 0.99\linewidth]{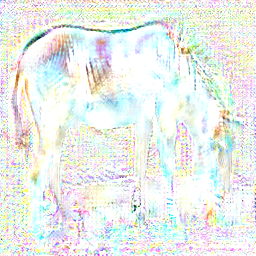}}
        \vskip -0.05in
        \textbf{\emph{\texttt{\fontsize{6.5pt}{\baselineskip}\selectfont -}}}
        \end{minipage}

		\begin{minipage}[b]{0.24\linewidth}
    		\centering
    		\includegraphics[width = 0.99\linewidth]{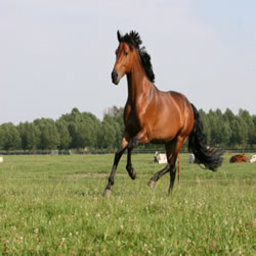}
    	\vskip -0.05in
        \textbf{\emph{\texttt{\fontsize{6.5pt}{\baselineskip}\selectfont FID:}}}
        \end{minipage}
    	\begin{minipage}[b]{0.24\linewidth}
    		\centering
    		\includegraphics[width = 0.99\linewidth]{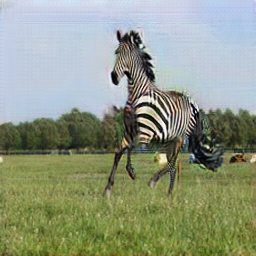}
    	\vskip -0.05in
        \textbf{\emph{\texttt{\fontsize{6.5pt}{\baselineskip}\selectfont 52.627}}}
        \end{minipage}
    	\begin{minipage}[b]{0.24\linewidth}
    		\centering
    		\includegraphics[width = 0.99\linewidth]{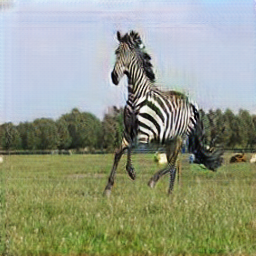}
    	\vskip -0.05in
        \textbf{\emph{\texttt{\fontsize{6.5pt}{\baselineskip}\selectfont 53.165}}}
        \end{minipage}
    	\begin{minipage}[b]{0.24\linewidth}
    		\centering
            \setlength{\fboxrule}{0.4pt}        \setlength{\fboxsep}{0cm}
    		\fbox{\includegraphics[width = 0.99\linewidth]{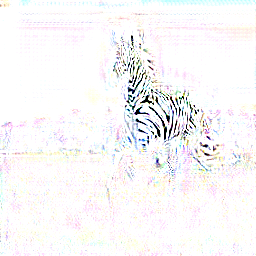}}
		\vskip -0.05in
        \textbf{\emph{\texttt{\fontsize{6.5pt}{\baselineskip}\selectfont -}}}
        \end{minipage}
\end{center}
	\caption{Representative image-to-image translation results: CycleGAN. Row 1: zebra to horse, Row 2: horse to zebra.}
	\label{Fig.8}
\end{figure}

\textbf{Image-to-image Translation}.
In addition to the StarGAN results above (Section~\ref{sec-naiveGANCompression}, Figure~\ref{Fig.1}), we apply the proposed compression method to CycleGAN~\citep{zhu2017unpaired} performing bidirectional translation between zebra and horse images. Results are shown in Figure~\ref{Fig.8}.

\textbf{Super Resolution}.
\begin{figure*}[!ht]
\begin{center}
    \centering
    \begin{minipage}[b]{0.195\linewidth}
		\centering
		\textbf{\texttt{\fontsize{5.5pt}{\baselineskip}\selectfont Ground Truth}}
	\end{minipage}
	\begin{minipage}[b]{0.195\linewidth}
		\centering
		\textbf{\texttt{\fontsize{5.5pt}{\baselineskip}\selectfont Dense Generator}}
	\end{minipage}
	\begin{minipage}[b]{0.195\linewidth}
		\centering
		\textbf{\texttt{\fontsize{5.5pt}{\baselineskip}\selectfont 50\% Filter-pruned Generator}}
	\end{minipage}
	\begin{minipage}[b]{0.195\linewidth}
		\centering
        \textbf{\texttt{\fontsize{5.5pt}{\baselineskip}\selectfont 50\% Fine-grained Generator}}
	\end{minipage}
	\begin{minipage}[b]{0.195\linewidth}
		\centering
        \textbf{\texttt{\fontsize{5.5pt}{\baselineskip}\selectfont 90\% Fine-grained Generator}}
	\end{minipage}

    \begin{minipage}[b]{0.195\linewidth}
		\centering
		\includegraphics[width = 0.81\linewidth]{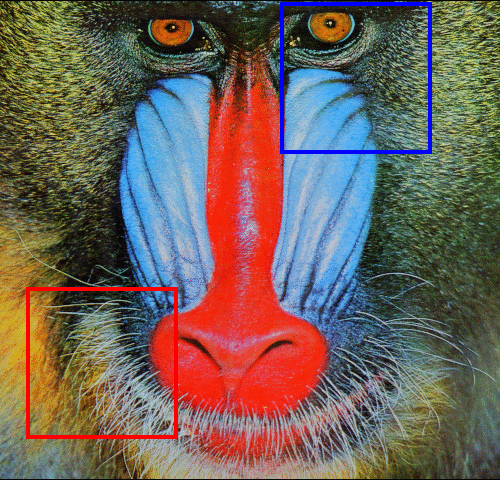}
	\end{minipage}
	\begin{minipage}[b]{0.195\linewidth}
		\centering
		\includegraphics[width = 0.81\linewidth]{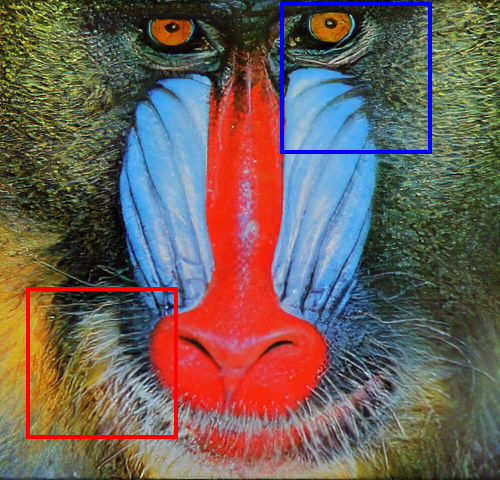}
	\end{minipage}
	\begin{minipage}[b]{0.195\linewidth}
		\centering
		\includegraphics[width = 0.81\linewidth]{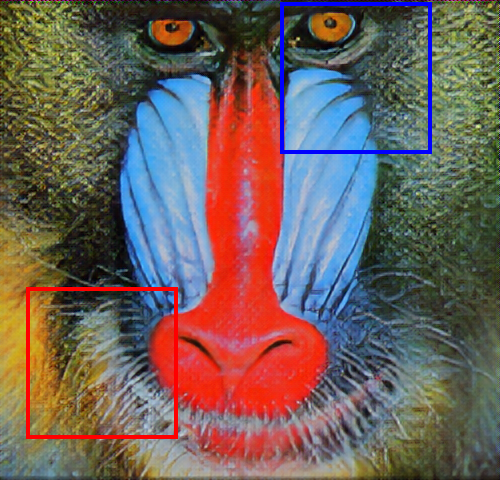}	
	\end{minipage}
	\begin{minipage}[b]{0.195\linewidth}
		\centering
		\includegraphics[width = 0.81\linewidth]{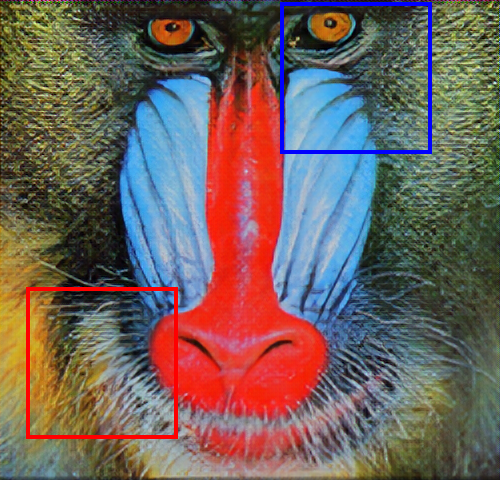}	
	\end{minipage}
    \begin{minipage}[b]{0.195\linewidth}
		\centering
		\includegraphics[width = 0.81\linewidth]{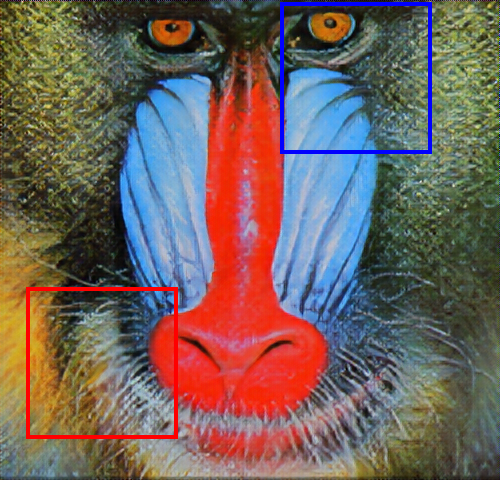}	
	\end{minipage}

    \begin{minipage}[b]{0.095\linewidth}
		\centering
		\includegraphics[width = 0.66\linewidth]{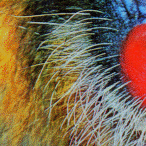}
	\end{minipage}
    \begin{minipage}[b]{0.095\linewidth}
		\centering
		\includegraphics[width = 0.66\linewidth]{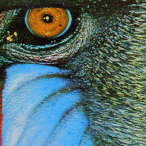}
	\end{minipage}
    \begin{minipage}[b]{0.095\linewidth}
		\centering
		\includegraphics[width = 0.66\linewidth]{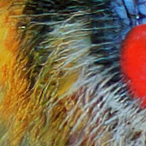}
	\end{minipage}
    \begin{minipage}[b]{0.095\linewidth}
		\centering
		\includegraphics[width = 0.66\linewidth]{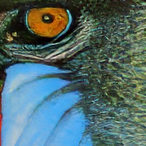}
	\end{minipage}
	\begin{minipage}[b]{0.095\linewidth}
		\centering
		\includegraphics[width = 0.66\linewidth]{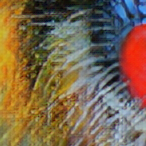}	
	\end{minipage}
	\begin{minipage}[b]{0.095\linewidth}
		\centering
		\includegraphics[width = 0.66\linewidth]{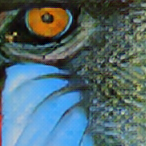}	
	\end{minipage}
	\begin{minipage}[b]{0.095\linewidth}
		\centering
		\includegraphics[width = 0.66\linewidth]{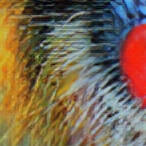}	
	\end{minipage}
	\begin{minipage}[b]{0.095\linewidth}
		\centering
		\includegraphics[width = 0.66\linewidth]{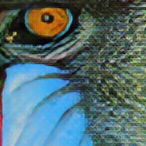}	
	\end{minipage}
	\begin{minipage}[b]{0.095\linewidth}
		\centering
		\includegraphics[width = 0.66\linewidth]{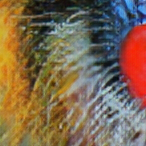}	
	\end{minipage}
	\begin{minipage}[b]{0.095\linewidth}
		\centering
		\includegraphics[width = 0.66\linewidth]{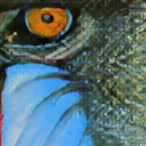}	
	\end{minipage}

\end{center}
	\caption{Representative super resolution results: SRGAN (with enlargements of boxed areas).}
	\label{Fig.11}
\end{figure*}
We apply self-supervised compression to SRGAN~\citep{ledig2017photo}\footnote{SRGAN baseline:
\url{https://github.com/xinntao/BasicSR}.}, which uses a discriminator network trained to differentiate between upscaled and the original high-resolution images. We trained SRGAN on the DIV2K data set~\cite{agustsson2017ntire}, and use the DIV2K validation images, as well as Set5~\cite{bevilacqua2012low} and Set14~\cite{zeyde2010single} to report deployment quality. In this task, quality is often evaluated by two metrics: Peak Signal-to-Noise Ratio (PSNR)~\citep{4550695} and Structural Similarity (SSIM)~\citep{wang2004image}. We also show FID scores~\citep{heusel2017gans} for our results in the results summarized in Table~\ref{Table.1}, and a representative output is shown in Figure~\ref{Fig.11}. These results also include filter-pruned generators (see Section~\ref{sec-pruningRatioGranularity}).

\begin{table}[htb]
\caption{\emph{PSNR}, \emph{SSIM} and \emph{FID} indicators for Validation Datasets}
\label{Table.1}
\centering
\resizebox{\linewidth}{!}{
\begin{tabular}{lccccccccc}
\toprule
\multirow{2}{*}{Dataset} & \multicolumn{3}{c}{Original Generator} & \multicolumn{3}{c}{Filter-Compressed G} & \multicolumn{3}{c}{Element-Compressed G}  \\
\cmidrule(lr){2-4}\cmidrule(lr){5-7}\cmidrule(lr){8-10}
                         & PSNR      & SSIM     & FID             & PSNR      & SSIM     & FID              & PSNR      & SSIM     & FID
                         \\
\midrule
Set5                     & 30.063 & 0.853 & 30.762       & 30.234 & 0.860 & 35.514        & 30.484 & 0.862 & 36.824                    \\
Set14                    & 26.643 & 0.716 & 55.457       & 27.315 & 0.745 & 82.118        & 27.417 & 0.744 & 70.126                     \\
DIV2K                    & 28.206 & 0.778 & 14.653       & 28.876 & 0.801 & 18.500        & 28.975 & 0.801 & 16.609
\\
\bottomrule
\end{tabular}
}
\end{table}


\section{Effect of Pruning Ratio and Granularity}\label{sec-pruningRatioGranularity}
After showing that self-supervised compression applies to many tasks and networks with a moderate, fine-grained sparsity of 50\%, we expand the scope of the investigation to include different pruning granularities and rates.  From coarse to fine, we can compress and remove the entire filters (3D-level), kernels (2D-level), vectors (1D-level) or individual elements (0D-level). In general, finer-grained pruning results in higher accuracy for a given sparsity rate, but coarser granularities are easier to exploit for performance gains due to their regular structure.
Similarly, different sparsity rates, leaving many nonzero weights or few, can result in varying levels of quality in the final network.

\begin{figure*}[htb]
\begin{center}
    \centering
    \begin{minipage}[b]{0.152\linewidth}
		\centering
		\textbf{\texttt{\fontsize{7.5pt}{\baselineskip}\selectfont Sparsity}}
    \end{minipage}
    \begin{minipage}[b]{0.152\linewidth}
		\centering
		\textbf{\texttt{\fontsize{7.5pt}{\baselineskip}\selectfont 0\%}}
    \end{minipage}
    \begin{minipage}[b]{0.152\linewidth}
		\centering
		\textbf{\texttt{\fontsize{7.5pt}{\baselineskip}\selectfont 25\%}}
    \end{minipage}
    \begin{minipage}[b]{0.152\linewidth}
		\centering
		\textbf{\texttt{\fontsize{7.5pt}{\baselineskip}\selectfont 50\%}}
    \end{minipage}
    \begin{minipage}[b]{0.152\linewidth}
		\centering
		\textbf{\texttt{\fontsize{7.5pt}{\baselineskip}\selectfont 75\%}}
    \end{minipage}
    \begin{minipage}[b]{0.152\linewidth}
		\centering
		\textbf{\texttt{\fontsize{7.5pt}{\baselineskip}\selectfont 90\%}}
    \end{minipage}

    \begin{minipage}[b]{0.152\linewidth}
		\centering
		\textbf{\emph{\texttt{\fontsize{7.5pt}{\baselineskip}\selectfont Fine-grained \\ ~ \\ ~ \\ FID:}}}
    \end{minipage}
    \begin{minipage}[b]{0.152\linewidth}
		\centering
		\includegraphics[width = 0.87\linewidth]{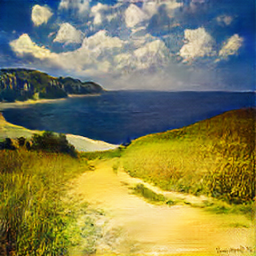}
    \vskip -0.05in
    \textbf{\emph{\texttt{\fontsize{7.5pt}{\baselineskip}\selectfont 32.006}}}
    \end{minipage}
    \begin{minipage}[b]{0.152\linewidth}
		\centering
		\includegraphics[width = 0.87\linewidth]{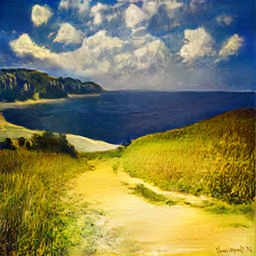}
    \vskip -0.05in
    \textbf{\emph{\texttt{\fontsize{7.5pt}{\baselineskip}\selectfont 32.462}}}
    \end{minipage}
    \begin{minipage}[b]{0.152\linewidth}
		\centering
		\includegraphics[width = 0.87\linewidth]{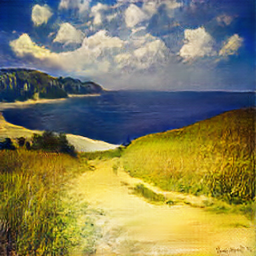}
    \vskip -0.05in
    \textbf{\emph{\texttt{\fontsize{7.5pt}{\baselineskip}\selectfont 33.387}}}
    \end{minipage}
    \begin{minipage}[b]{0.152\linewidth}
		\centering
		\includegraphics[width = 0.87\linewidth]{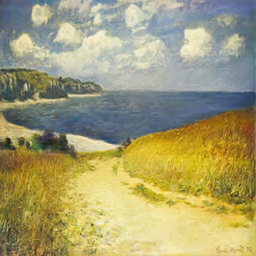}
    \vskip -0.05in
    \textbf{\emph{\texttt{\fontsize{7.5pt}{\baselineskip}\selectfont 34.543}}}
    \end{minipage}
    \begin{minipage}[b]{0.152\linewidth}
		\centering
		\includegraphics[width = 0.87\linewidth]{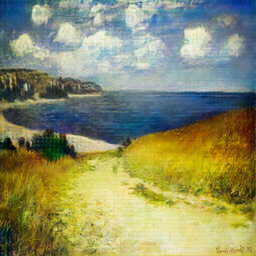}
    \vskip -0.05in
    \textbf{\emph{\texttt{\fontsize{7.5pt}{\baselineskip}\selectfont 41.251}}}
    \end{minipage}

    \begin{minipage}[b]{0.152\linewidth}
		\centering
		\textbf{\emph{\texttt{\fontsize{7.5pt}{\baselineskip}\selectfont Filter-pruned \\ ~ \\ ~ \\ FID:}}}
    \end{minipage}
    \begin{minipage}[b]{0.152\linewidth}
		\centering
		\includegraphics[width = 0.87\linewidth]{monet2photo_cyclegan_00010_fake_B.png}
    \vskip -0.05in
    \textbf{\emph{\texttt{\fontsize{7.5pt}{\baselineskip}\selectfont 32.006}}}
    \end{minipage}
    \begin{minipage}[b]{0.152\linewidth}
		\centering
		\includegraphics[width = 0.87\linewidth]{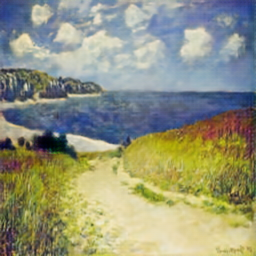}
    \vskip -0.05in
    \textbf{\emph{\texttt{\fontsize{7.5pt}{\baselineskip}\selectfont 82.349}}}
    \end{minipage}
    \begin{minipage}[b]{0.152\linewidth}
		\centering
		\includegraphics[width = 0.87\linewidth]{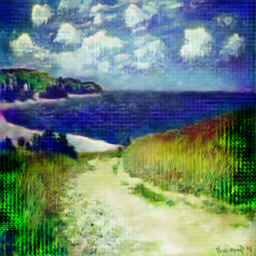}
    \vskip -0.05in
    \textbf{\emph{\texttt{\fontsize{7.5pt}{\baselineskip}\selectfont 105.884}}}
    \end{minipage}
    \begin{minipage}[b]{0.152\linewidth}
		\centering
		\includegraphics[width = 0.87\linewidth]{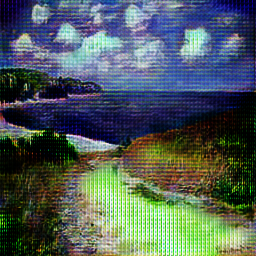}
    \vskip -0.05in
    \textbf{\emph{\texttt{\fontsize{7.5pt}{\baselineskip}\selectfont 182.277}}}
    \end{minipage}
    \begin{minipage}[b]{0.152\linewidth}
		\centering
		\includegraphics[width = 0.87\linewidth]{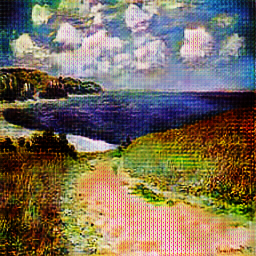}
    \vskip -0.05in
    \textbf{\emph{\texttt{\fontsize{7.5pt}{\baselineskip}\selectfont 204.795}}}
    \end{minipage}
\end{center}
	\caption{Representative results for pruning rate and granularity study of style transfer.}
	\label{Fig.ratiogranularity}
\end{figure*}

\begin{figure*}[!htb]
\begin{center}
    \begin{minipage}[b]{0.015\linewidth}
		\centering
		\textbf{\texttt{\fontsize{6.5pt}{\baselineskip}\selectfont Sparsity}}
    \end{minipage}
    \begin{minipage}[b]{0.47\linewidth}
		\centering
		\textbf{\texttt{\fontsize{6.5pt}{\baselineskip}\selectfont Fine-grained}}
    \end{minipage}
    \begin{minipage}[b]{0.015\linewidth}
		\centering
		\textbf{\texttt{\fontsize{6.5pt}{\baselineskip}\selectfont Sparsity}}
    \end{minipage}
    \begin{minipage}[b]{0.47\linewidth}
		\centering
		\textbf{\texttt{\fontsize{6.5pt}{\baselineskip}\selectfont Filter-pruned}}
    \end{minipage}

    \begin{minipage}[b]{0.035\linewidth}
		\centering
		\textbf{\texttt{\fontsize{5.5pt}{\baselineskip}\selectfont 0\% \\ ~ \\ }}
    \end{minipage}
    \begin{minipage}[b]{0.45\linewidth}
		\centering
		\includegraphics[width = 0.89\linewidth]{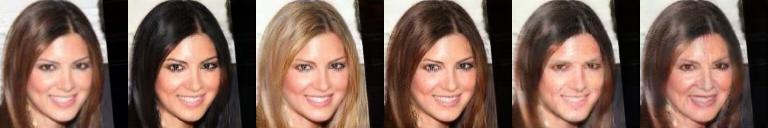}	
	\end{minipage}
    \begin{minipage}[b]{0.035\linewidth}
		\centering
		\textbf{\texttt{\fontsize{5.5pt}{\baselineskip}\selectfont 0\% \\ ~ \\ }}
    \end{minipage}
	\begin{minipage}[b]{0.45\linewidth}
		\centering
		\includegraphics[width = 0.89\linewidth]{stargan_celeba_128_101-images.jpg}	
	\end{minipage}

    \begin{minipage}[b]{0.035\linewidth}
		\centering
		\textbf{\texttt{\fontsize{5.5pt}{\baselineskip}\selectfont 25\% \\ ~ \\ }}
    \end{minipage}
	\begin{minipage}[b]{0.45\linewidth}
		\centering
		\includegraphics[width = 0.89\linewidth]{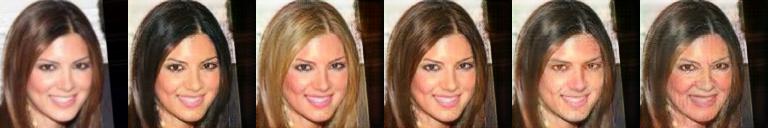}	
	\end{minipage}
    \begin{minipage}[b]{0.035\linewidth}
		\centering
		\textbf{\texttt{\fontsize{5.5pt}{\baselineskip}\selectfont 25\% \\ ~ \\ }}
    \end{minipage}
    \begin{minipage}[b]{0.45\linewidth}
		\centering
		\includegraphics[width = 0.89\linewidth]{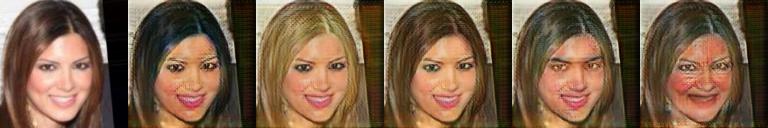}
	\end{minipage}

    \begin{minipage}[b]{0.035\linewidth}
		\centering
		\textbf{\texttt{\fontsize{5.5pt}{\baselineskip}\selectfont 50\% \\ ~ \\ }}
    \end{minipage}
	\begin{minipage}[b]{0.45\linewidth}
		\centering
		\includegraphics[width = 0.89\linewidth]{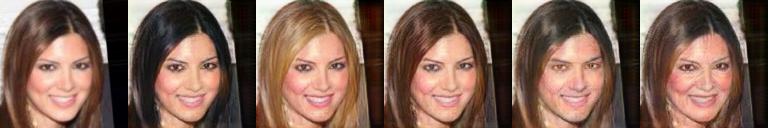}	
	\end{minipage}
    \begin{minipage}[b]{0.035\linewidth}
		\centering
		\textbf{\texttt{\fontsize{5.5pt}{\baselineskip}\selectfont 50\% \\ ~ \\ }}
    \end{minipage}
    \begin{minipage}[b]{0.45\linewidth}
		\centering
		\includegraphics[width = 0.89\linewidth]{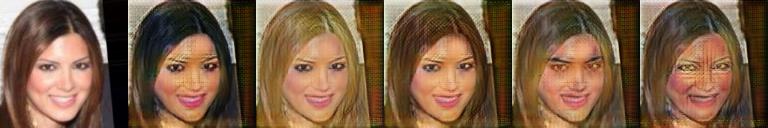}
	\end{minipage}

    \begin{minipage}[b]{0.035\linewidth}
		\centering
		\textbf{\texttt{\fontsize{5.5pt}{\baselineskip}\selectfont 75\% \\ ~ \\ }}
    \end{minipage}
    \begin{minipage}[b]{0.45\linewidth}
		\centering
		\includegraphics[width = 0.89\linewidth]{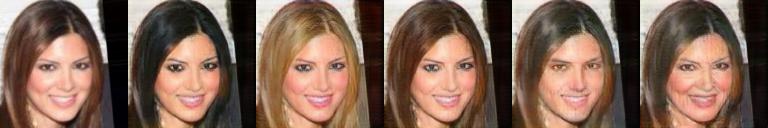}
	\end{minipage}
    \begin{minipage}[b]{0.035\linewidth}
		\centering
		\textbf{\texttt{\fontsize{5.5pt}{\baselineskip}\selectfont 75\% \\ ~ \\ }}
    \end{minipage}
    \begin{minipage}[b]{0.45\linewidth}
		\centering
		\includegraphics[width = 0.89\linewidth]{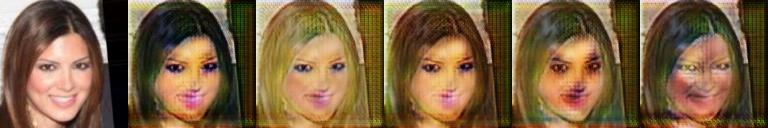}
	\end{minipage}

    \begin{minipage}[b]{0.035\linewidth}
		\centering
		\textbf{\texttt{\fontsize{5.5pt}{\baselineskip}\selectfont 90\% \\ ~ \\ }}
    \end{minipage}
    \begin{minipage}[b]{0.45\linewidth}
		\centering
		\includegraphics[width = 0.89\linewidth]{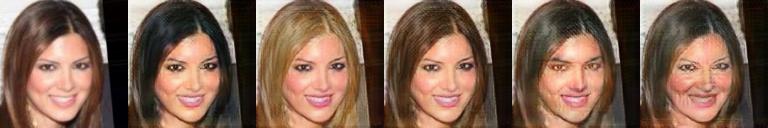}
	\end{minipage}
    \begin{minipage}[b]{0.035\linewidth}
		\centering
		\textbf{\texttt{\fontsize{5.5pt}{\baselineskip}\selectfont 90\% \\ ~ \\ }}
    \end{minipage}
    \begin{minipage}[b]{0.45\linewidth}
		\centering
		\includegraphics[width = 0.89\linewidth]{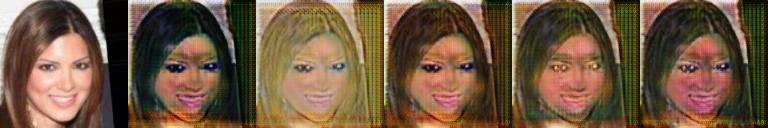}
	\end{minipage}
\end{center}
	\caption{Representative results for pruning rate and granularity study of image-to-image translation.}
	\label{Fig.ratiogranularity_StarGAN}
\end{figure*}

We pruned all tasks by removing both single elements (0D) and entire filters (3D). Further, for each granularity, we pruned to final sparsities of 25\%, 50\%, 75\%, and 90\%. Representative results for CycleGAN (Monet $\rightarrow$ Photo) and StarGAN are shown in Figure~\ref{Fig.ratiogranularity} and Figure~\ref{Fig.ratiogranularity_StarGAN}, with results for all tasks in the \emph{\textbf{Appendix}}. In general, 0D pruning is less invasive, even at higher sparsities. Up to 90\% fine-grained sparsity, some fine details faded away in CycleGAN and StarGAN, but filter pruning results in drastic color shifts and loss of details at even 25\% sparsity.

\section{Conclusion and Future Work}
Network pruning has been applied to many tasks, but never to GANs performing complex tasks. We showed that existing pruning approaches fail to retain network quality, as do training modifications aimed at compressing simple GANs by other methods applied to pruning. To solve this, we used a pre-trained discriminator to self-supervise the pruning of several GANs' generators and showed this method performs well both qualitatively and quantitatively.  Advantages of our method include:

\begin{itemize}
\vspace{-0.5em}
    \item The results from the compressed generators are greatly improved over past work.
    \item The self-supervised compression is much shorter than the original GAN training process - only 1-10\% of the original training time is needed.
    \item It is an end-to-end compression schedule that does not require objective evaluation metrics; final quality is accurately reflected in loss curves.
    \item We introduce a single optional hyperparameter (fixed to 0.5 for all our experiments).
\vspace{-0.5em}
\end{itemize}

We use self-supervised GAN compression to show that pruning whole filters, which can work well for image classification models~\citep{li2016pruning}, may perform poorly for GAN applications. Even pruned at a moderate sparsity (e.g. 25\% in Figure~\ref{Fig.ratiogranularity}), the generated image has an obvious color shift and does not transfer the photorealistic style. In contrast, the fine-grained compression stategy works well for all tasks we explored. SRGAN seems to be an exception to filter-pruning's poor results; we have to look closely to see differences, and it's not clear which is subjectively better.

We have not tried to achieve extremely aggressive compression rates with complicated pruning strategies. Different models may be able to tolerate different amounts of pruning when applied to a task, which we leave to future work. Similarly, we have used network pruning to show the importance and utility of the proposed method, but self-supervised compression is general to other techniques, such as quantization, weight sharing, etc. There are other tasks for which GANs can provide compelling results, and newer networks for tasks we have already explored; future work will extend our self-supervised compression method to these new areas. Finally, self-supervised compression may apply to other network types and tasks if a discriminator is trained alongside the teacher and student networks.

\section{Appendix}
More experiments and details can refer to the appendix file in repository: \url{https://gitlab.com/dxxz/Self-Supervised-GAN-Compression-Appendix}. 

\clearpage

\bibliographystyle{icml2020}
\bibliography{icml2020}

\end{document}